\documentclass[letterpaper, 10 pt, conference]{ieeeconf}

\IEEEoverridecommandlockouts
\usepackage{cite}
\usepackage{amsmath,amssymb,amsfonts}
\usepackage{graphicx}
\usepackage{textcomp}
\usepackage{xcolor}
\usepackage{booktabs}
\usepackage{multirow}
\usepackage{url}
\usepackage[caption=false,font=footnotesize]{subfig} 

\usepackage{array}
\usepackage{microtype}

\title{\LARGE \bf
MPTF-Net: Multi-view Pyramid Transformer Fusion Network for \\ LiDAR-based Place Recognition
}

\author{
Shuyuan Li,
Zihang Wang,
Xieyuanli Chen,
Wenkai Zhu,
Xiaoteng Fang, \\ 
Peizhou Ni,
Junhao Yang,
and Dong Kong
}

\begin{document}

\maketitle
\thispagestyle{empty}
\pagestyle{empty}

\begin{abstract}

LiDAR-based place recognition (LPR) is essential for global localization and loop-closure detection in large-scale SLAM systems. Existing methods typically construct global descriptors from Range Images or BEV representations for matching. BEV is widely adopted due to its explicit 2D spatial layout encoding and efficient retrieval. However, conventional BEV representations rely on simple statistical aggregation, which fails to capture fine-grained geometric structures, leading to performance degradation in complex or repetitive environments. To address this, we propose MPTF-Net, a novel multi-view multi-scale pyramid Transformer fusion network. Our core contribution is a multi-channel NDT-based BEV encoding that explicitly models local geometric complexity and intensity distributions via Normal Distribution Transform, providing a noise-resilient structural prior. To effectively integrate these features, we develop a customized pyramid Transformer module that captures cross-view interactive correlations between Range Image Views (RIV) and NDT-BEV at multiple spatial scales. Extensive experiments on the nuScenes, KITTI and NCLT datasets demonstrate that MPTF-Net achieves state-of-the-art performance, specifically attaining a Recall@1 of 96.31\% on the nuScenes Boston split while maintaining an inference latency of only 10.02 ms, making it highly suitable for real-time autonomous unmanned systems.

\end{abstract}

\section{INTRODUCTION}
\label{sec:intro}

Reliable long-term autonomy is a fundamental prerequisite for deploying intelligent vehicles and autonomous unmanned systems in complex, large-scale environments \cite{pseduo}. Within the Simultaneous Localization and Mapping (SLAM) framework, place recognition plays a pivotal role in mitigating accumulated drift and maintaining global map consistency. Although vision-based approaches have achieved remarkable progress, their performance remains inherently vulnerable to drastic illumination variations and adverse weather conditions. In contrast, LiDAR-based Place Recognition (LPR), which provides stable and active $360^{\circ}$ geometric perception, has emerged as a critical safeguard for robust global localization.

Despite substantial advances, achieving highly discriminative and robust LPR in large-scale or structurally repetitive urban environments remains challenging. To enable efficient processing of unstructured 3D point clouds, most existing methods project raw LiDAR scans into 2D representations, such as Range Image Views (RIV) or Bird's Eye Views (BEV). However, single-view projections inevitably induce information loss. RIV representations are sensitive to severe scale variations and occlusions, while conventional BEV generation typically relies on low-order statistical aggregations, such as maximum height or binary occupancy. These fundamentally first-order abstractions discard local covariance structures and intensity distributions, thereby limiting discriminative capacity in geometrically complex scenes \cite{ralbev}. Furthermore, although multi-view strategies attempt to alleviate single-view limitations, effectively modeling fine-grained cross-view interactions across spatial scales---without incurring prohibitive computational overhead---remains largely underexplored.

\begin{figure}[!t] 
    \centering
    \includegraphics[width=0.5\textwidth]{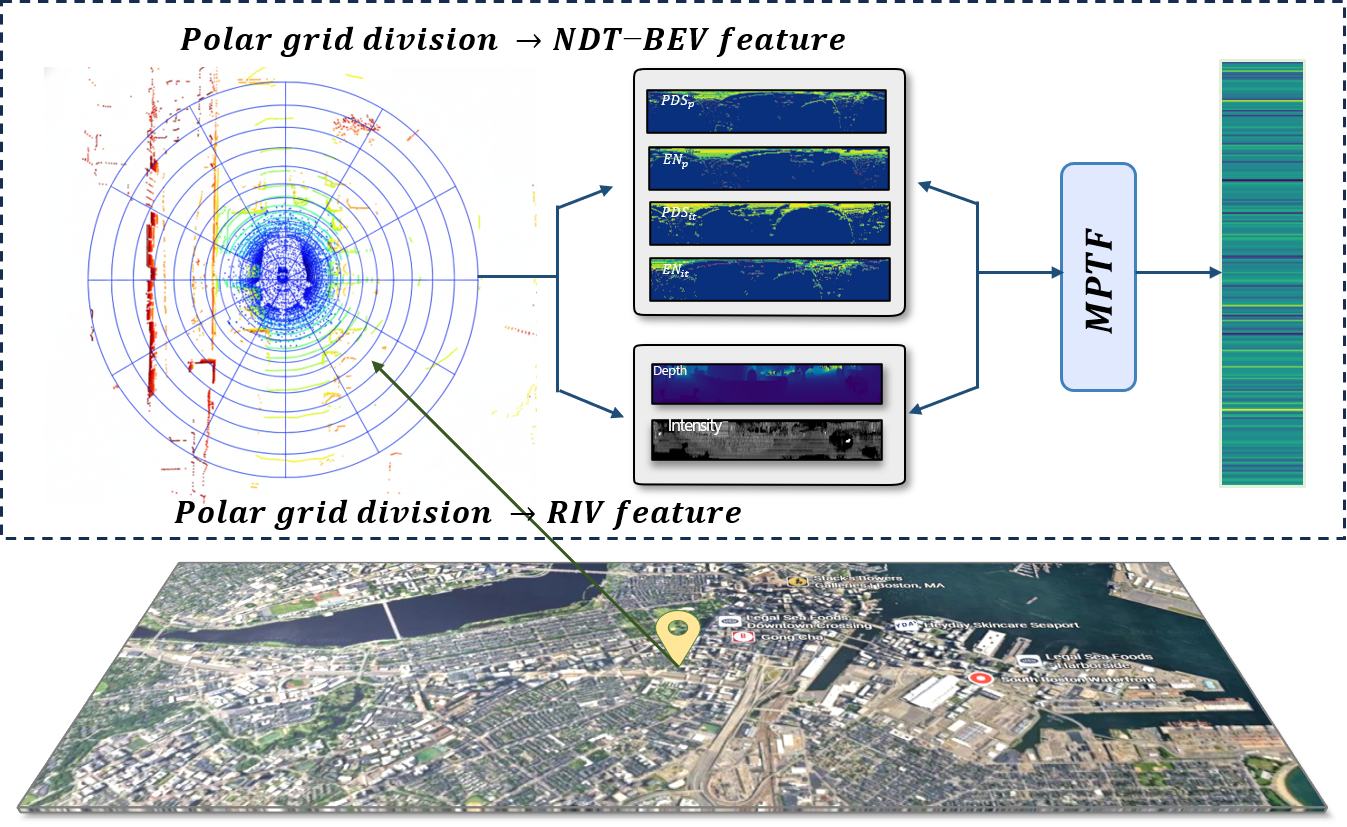}
    \caption{Overview of the proposed MPTF-Net, a novel multi-view fusion-driven global descriptor extraction network for LiDAR-based place recognition.}
    \label{fig:overview}
\end{figure}

To address these challenges, we propose a novel Multi-view Pyramid Transformer Fusion Network (MPTF-Net), as illustrated in Fig.~\ref{fig:overview}. Rather than relying on simplistic occupancy-based encoding, our approach introduces a multi-channel BEV representation grounded in the Normal Distribution Transform (NDT) \cite{cctnet}. By explicitly modeling local point clusters as multivariate probability density functions, the proposed NDT-BEV captures second-order geometric statistics and intensity uncertainty, yielding a discriminative and noise-resilient structural prior. To fully exploit the complementary nature of RIV and BEV projections, we further design a customized Multi-scale Pyramid Transformer Fusion (MPTF) module. This module employs hierarchically aligned bi-directional cross-attention to capture latent inter-view correlations across multiple spatial resolutions \cite{pointnetvlad, ndt}. Importantly, by omitting absolute positional embeddings, the network preserves intrinsic robustness to viewpoint shifts and axial yaw rotations. Finally, we incorporate a context-gated enhanced NetVLAD \cite{netvlad} module to perform adaptive and selective global feature aggregation \cite{scan}. 

Extensive evaluations on the nuScenes, NCLT, and KITTI benchmarks demonstrate consistent state-of-the-art performance across diverse environments. In particular, MPTF-Net achieves a Recall@1 of 96.31\% on the Boston split and 99.43\% on the unseen Singapore split of nuScenes. Notably, the network maintains a real-time inference latency of 10.02 ms (100 Hz), highlighting its practical suitability for onboard deployment in autonomous unmanned systems.
The main contributions of this work are summarized as follows:
\begin{itemize}
    \item We propose a novel multi-view fusion-driven global descriptor extraction network tailored for LiDAR-based place recognition, which jointly encodes multi-scale geometric and intensity cues from RIV and BEV representations. We specifically leverage the Normal Distribution Transform (NDT) to generate BEV features that provide a noise-resilient structural prior, significantly enhancing discriminative power in complex urban environments.
    \item We design a Multi-scale Pyramid Transformer Fusion (MPTF) module to hierarchically aggregate features across spatial scales using a bi-directional cross-attention mechanism, capturing latent correlations between the RIV and BEV branches. This is coupled with a context-gating enhanced NetVLAD module to enable adaptive weighting and robust aggregation of multi-view descriptors.
    \item Extensive experiments on the nuScenes, KITTI and NCLT datasets demonstrate that MPTF-Net achieves state-of-the-art performance with exceptional zero-shot generalization, attaining a Recall@1 of 96.31\% on the Boston split and 99.43\% on the unseen Singapore split. Furthermore, the network maintains a real-time retrieval latency of 19.62 ms (50 Hz), ensuring its suitability for onboard deployment in autonomous unmanned systems.
\end{itemize}

\section{RELATED WORKS}\label{sec:formatting}
\subsection{Projection Image-Based LPR}

To reduce the computational burden of directly processing raw 3D point clouds, projecting LiDAR scans into compact 2D representations has become a widely adopted paradigm for large-scale place recognition~\cite{pr}. Kim and Kim~\cite{scan} introduced Scan Context (SC), a handcrafted global descriptor that encodes vertical structural information in a polar grid. Building upon this idea, Wang et al.~\cite{iris} proposed LiDAR-Iris, which employs Log-Gabor filtering to generate discriminative binary signature maps, while Zhou et al.~\cite{ndd} developed NDD, leveraging normal-distribution densities for efficient retrieval.

With the advancement of deep learning, projection-based learned descriptors have achieved significant progress. Luo et al.~\cite{bvmatch} presented BVMatch, which extracts rotation-invariant features from Bird's Eye View (BEV) images, and BEVPlace~\cite{bevplace} further enhanced viewpoint robustness using group convolutions. More recently, Wang et al.~\cite{global} proposed a probabilistic occupancy grid-based framework that achieves full Roll-Pitch-Yaw (RPY) invariance, significantly improving robustness under 6-DoF viewpoint changes. In addition, learning-based architectures such as OverlapNet~\cite{overlapnet} and OverlapTransformer~\cite{former} demonstrate the effectiveness of attention mechanisms for handling severe viewpoint variations.

\subsection{Multi-View-Based LPR}

To exploit complementary geometric cues from different spatial projections, multi-view LiDAR place recognition (LPR) frameworks have attracted increasing attention. Yin et al.~\cite{fusionvald} pioneered this direction with FusionVLAD, which integrates spherical and top-down projections at the descriptor level to enhance viewpoint robustness. To enable more explicit inter-view interaction, Ma et al.~\cite{cvt} proposed CVTNet, introducing a cross-view Transformer to model token-level dependencies between Range Image View (RIV) and BEV representations. More recently, Luo et al.~\cite{mrmt} proposed MRMT-PR, a multi-scale reverse-view architecture designed to improve robustness under extreme viewpoint variations.

Nevertheless, existing multi-view frameworks exhibit two primary limitations. First, most BEV-based pipelines rely on simplistic statistical aggregation (e.g., maximum height or binary occupancy) to construct feature maps. Such low-order representations inevitably discard fine-grained geometric details and intensity distributions, limiting discriminability in complex environments~\cite{ndttransformer}. Second, current fusion mechanisms often lack explicit hierarchical alignment. They typically operate at a single representation scale without modeling the cross-scale relationships between local texture details and high-level structural context. These deficiencies motivate the development of a framework that combines robust probabilistic BEV modeling with hierarchically aligned cross-view fusion.
\begin{figure*}[t]
  \centering
  \includegraphics[width=\linewidth]{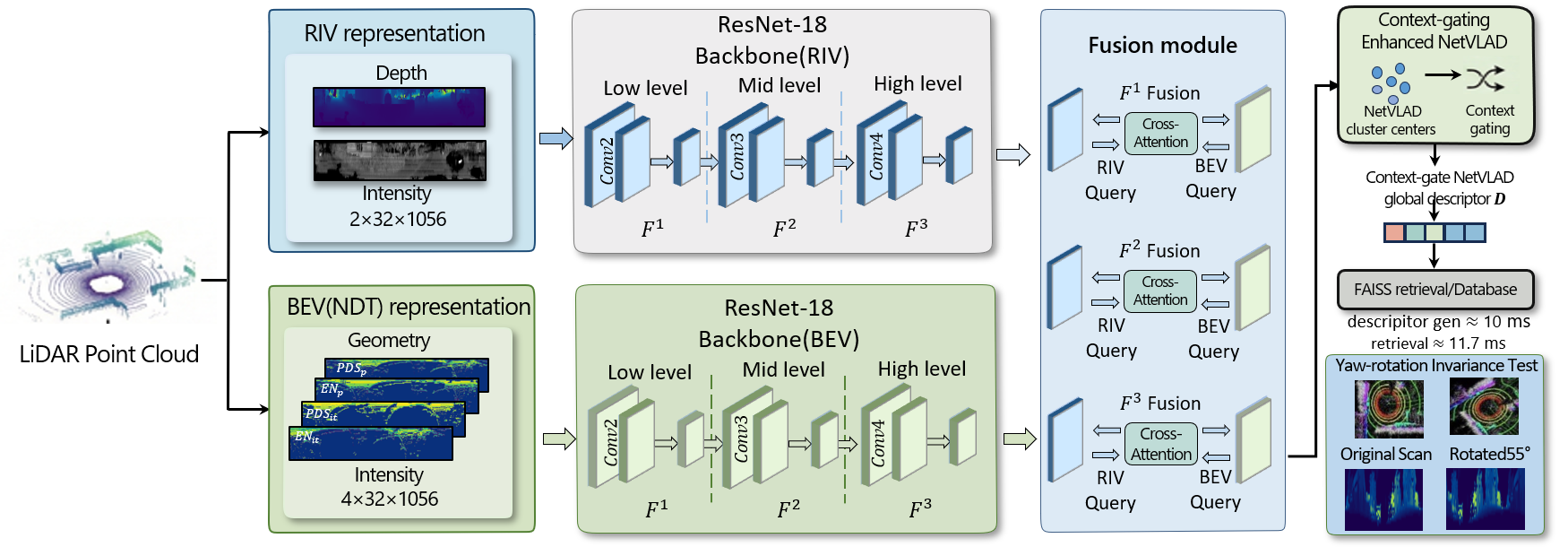}
  \caption{Overall pipeline of MPTF-Net. The network jointly exploits RIV and BEV representations containing geometric and intensity cues. RIV and BEV branches adopt ResNet-based backbones, and the multi-scale Transformer fusion module captures cross-view interactions. Finally, the context-gating enhanced NetVLAD aggregates the fused features into discriminative, viewpoint-invariant global descriptors.}
  \label{fig:framework}

\end{figure*}

\section{PROPOSED METHODOLOGY}

As shown in Fig.~\ref{fig:framework}, MPTF-Net extracts multi-scale features from Range Image View (RIV) and NDT-based Bird's Eye View (BEV) via parallel ResNet backbones. These features are fused by a pyramid Transformer with bi-directional cross-attention and finally aggregated by a Context-gating Enhanced NetVLAD to generate a discriminative global descriptor.

\subsection{RIV Dual-Feature Representation}

The Range Image View (RIV) provides a dense, ego-centric projection that preserves radial distance and angular distribution, retaining detailed near-field geometric structures. However, single-channel depth maps offer limited discriminative capability. To enhance feature expressiveness, we project 3D points onto a $32 \times 1056$ spherical grid and jointly encode normalized radial distance (0–80 m) and normalized backscatter intensity, forming a dual-channel representation that captures both spatial occupancy and radiometric properties.During projection, a depth-priority filling strategy is adopted to alleviate occlusion effects, ensuring that nearer points overwrite farther ones along the same ray. The resulting compact dual-channel tensor provides an informative and geometrically consistent input for subsequent multi-view fusion, effectively complementing the structural characteristics of BEV representations.


\subsection{BEV Multi-Feature Representation}


Conventional BEV encodings typically rely on simple statistical methods, such as occupancy counts or height pooling, which lose higher-order geometric structures and are often sensitive to sensor noise, sparse data, and isolated outliers. To address this limitation, we adopt the normal distribution transform (NDT) formulation, explicitly modeling the local spatial distribution within each BEV cell. By fitting a local Gaussian distribution, NDT naturally reduces the statistical weight of isolated outliers, providing more robust structural features that effectively mitigate noise interference in the representation.

To maintain azimuthal consistency with the RIV representation, the BEV space is discretized in polar coordinates. For a given polar grid cell containing a point cluster $\mathcal{P} = \{\mathbf{p}_1, \dots, \mathbf{p}_N\}$, the local geometry is modeled as a gaussian distribution parameterized by its mean $\boldsymbol{\mu}$ and covariance matrix $\boldsymbol{\Sigma}$:
\begin{equation}
\boldsymbol{\mu} = \frac{1}{N} \sum_{j=1}^{N} \mathbf{p}_j, \quad
\boldsymbol{\Sigma} = \frac{1}{N-1} \sum_{j=1}^{N} (\mathbf{p}_j - \boldsymbol{\mu})(\mathbf{p}_j - \boldsymbol{\mu})^T.
\label{eq:ndt_stats}
\end{equation}

Unlike first-order aggregation, the covariance matrix captures anisotropic surface characteristics and local structural variability, enabling the representation to distinguish planar, linear, and volumetric patterns. Based on this probabilistic modeling, we derive complementary statistics to characterize both geometric complexity and local point concentration.

The structural variability within each cell is quantified using the differential entropy of the Gaussian distribution:
\begin{equation}
\mathcal{H} = \frac{1}{2} \ln \left( (2\pi e)^D |\boldsymbol{\Sigma}| \right),
\label{eq:entropy}
\end{equation}
where $D=3$ denotes the spatial dimensionality. Regions with larger entropy correspond to geometrically complex or cluttered structures, which typically provide stronger discriminative cues for place recognition.

To further measure how tightly the observed points conform to the estimated distribution, we compute a probability density score (PDS),:
\begin{equation}
\text{PDS} = \sum_{j=1}^{N} 
\exp \left(
-\frac{1}{2} 
(\mathbf{p}_j - \boldsymbol{\mu})^T 
\boldsymbol{\Sigma}^{-1} 
(\mathbf{p}_j - \boldsymbol{\mu})
\right),
\label{eq:grs}
\end{equation}
where the normalization constant is omitted since only relative responses are required. This statistic reflects the concentration of points under the learned local distribution and provides a density-aware structural descriptor that is less sensitive to raw point counts.

In addition to spatial coordinates, reflectance intensity values are modeled independently within each cell using a one-dimensional Gaussian distribution, from which entropy and Gaussian response statistics are computed analogously. This enables the representation to jointly encode geometric and radiometric characteristics.
\begin{figure}[t]
  \centering
  \includegraphics[keepaspectratio,width=1\linewidth]{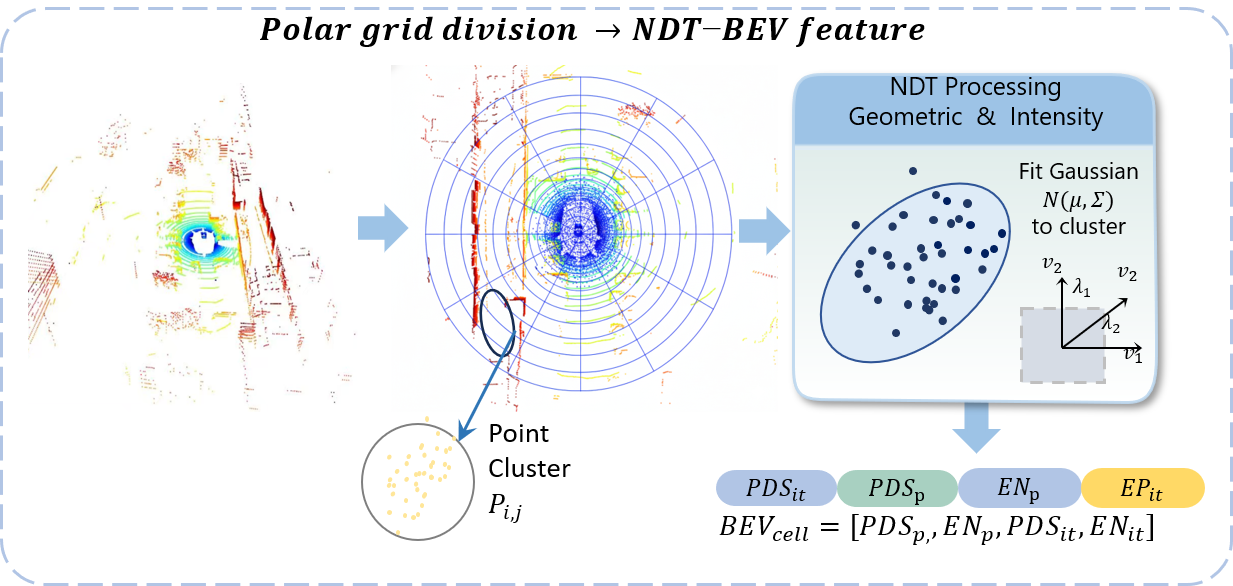}
  \caption{Block diagram of the BEV multi-feature encoding structure. After dividing the polar coordinate grids and selecting the point cloud clusters within the grids, NDT methods are utilized to compute geometric and intensity statistics.}
  \label{fig:bev_structure}
\end{figure}

The final BEV feature map is constructed as
\begin{equation}
\mathbf{P}_{bev} =
\left[
\text{PDS}_p,\;
\text{EN}_p,\;
\text{PDS}_{it},\;
\text{EN}_{it}
\right],
\label{eq:bev_feature}
\end{equation}
where subscripts $p$ and $it$ denote spatial position and intensity, respectively.

By explicitly incorporating second-order statistics in both geometric and intensity domains, the proposed representation preserves fine-grained structural anisotropy that is typically lost in conventional occupancy-based BEV encoding. This probabilistic modeling provides a noise-resilient and discriminative structural prior, significantly enhancing robustness under viewpoint variations and environmental changes.

\begin{figure}[t]
\centering

\subfloat[Geometric Entropy]{%
  \includegraphics[width=0.48\linewidth]{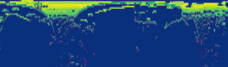}
}
\hfill
\subfloat[Intensity Entropy]{%
  \includegraphics[width=0.48\linewidth]{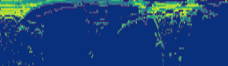}
}

\vspace{2mm}

\subfloat[Geometric PDS ($\mathrm{PDS}_p$)]{%
  \includegraphics[width=0.48\linewidth]{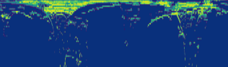}
}
\hfill
\subfloat[Intensity PDS ($\mathrm{PDS}_{it}$)]{%
  \includegraphics[width=0.48\linewidth]{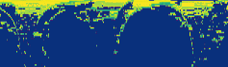}
}

\caption{Visualization of multimodal BEV features. These maps capture complementary structural and radiometric information.}
\label{fig:bev_features}

\end{figure}

\subsection{Multi-scale Cross-view Fusion}

To effectively integrate the complementary characteristics of RIV (fine-grained elevation textures) and BEV (global planar structures), we propose an Azimuth-Aligned Multi-scale Cross-view Fusion module.

Given that both RIV and BEV originate from the same LiDAR scan, they share an identical discretization along the azimuth dimension. Let $\mathbf{F}_r^i \in \mathbb{R}^{H_r^i \times W \times C}$ and $\mathbf{F}_b^i \in \mathbb{R}^{H_b^i \times W \times C}$ denote the RIV and BEV features at pyramid level $i$, respectively, where $W$ corresponds to the shared azimuth bins. Although the vertical dimensions ($H_r^i$ representing elevation and $H_b^i$ representing range) encode different physical meanings, each column indexed by $w \in \{1, \dots, W\}$ represents observations from the exact same angular direction in 3D space.

This strict azimuthal alignment provides a natural geometric prior for cross-view interaction. Instead of performing unconstrained global attention which incurs high computational cost and ignores geometric constraints, we explicitly enforce alignment-aware feature association.

For each scale $i$, we first project the features into a unified embedding space:
\begin{equation}
\mathbf{Q}_r^i = \phi_q(\mathbf{F}_r^i), \quad
\mathbf{K}_b^i = \phi_k(\mathbf{F}_b^i), \quad
\mathbf{V}_b^i = \phi_v(\mathbf{F}_b^i)
\label{eq:projection}
\end{equation}
where $\phi_q, \phi_k, \phi_v$ denote learnable linear projections implemented via $1\times1$ convolutions.

To preserve geometric consistency, attention is computed strictly within the aligned azimuth bins. For the $w$-th azimuth bin, the interaction is formulated as:
\begin{equation}
\mathbf{A}_r^i(w) =
\text{softmax}\left(
\frac{\mathbf{Q}_r^i(w) (\mathbf{K}_b^i(w))^\top}{\sqrt{d_k}}
\right)\mathbf{V}_b^i(w)
\label{eq:aligned_attn}
\end{equation}
Here, $\mathbf{Q}_r^i(w) \in \mathbb{R}^{H_r^i \times C}$ and $\mathbf{K}_b^i(w) \in \mathbb{R}^{H_b^i \times C}$ are column vectors corresponding to the specific angle $w$. This operation effectively captures correlations between range intervals and elevation intervals within the same angular sector.

The fused features are further refined via a residual Feed-Forward Network (FFN):
\begin{equation}
\tilde{\mathbf{F}}_r^i = 
\mathbf{F}_r^i + \text{FFN}(\mathbf{A}_r^i)
\label{eq:refine}
\end{equation}
A symmetric operation is applied to update the BEV branch features. By enforcing alignment-aware interaction at each pyramid level, the proposed mechanism ensures that fine-grained local textures in RIV are consistently associated with structurally meaningful BEV representations. This hierarchical and geometrically constrained interaction enables robust descriptor learning under large viewpoint and environmental variations.

\subsection{Yaw-Rotation Invariance Analysis}
\label{sec:yaw_invariance}

Robust place recognition requires invariance to the vehicle's yaw angle. MPTF-Net achieves this through shift-equivariant feature extraction and multi-view fusion, followed by shift-invariant global aggregation.

Let $\mathbf{X} \in \mathbb{R}^{C \times H \times W}$ denote a feature tensor, where the width dimension corresponds to the discretized azimuth. A yaw rotation $\theta = \frac{2\pi k}{W}$ is equivalent to a cyclic shift operator $T_k$ along this dimension. With circular padding, the convolutional backbone $f(\cdot)$ preserves translational equivariance:
\begin{equation}
f(T_k \mathbf{X}) = T_k f(\mathbf{X}).
\end{equation}

The Azimuth-Aligned Fusion module maintains this property. Since the cross-attention operator $\mathcal{A}(\cdot)$ processes azimuth bins with shared weights, it satisfies permutation-equivariance:
\begin{equation}
\mathcal{A}(T_k \mathbf{Q}, T_k \mathbf{K}, T_k \mathbf{V}) 
= T_k \mathcal{A}(\mathbf{Q}, \mathbf{K}, \mathbf{V}).
\end{equation}

Shift-equivariance is converted to shift-invariance by the context-gated NetVLAD layer. As NetVLAD aggregates local descriptors via commutative spatial summation,
\begin{equation}
\mathcal{G}(T_k \mathbf{X}) = \mathcal{G}(\mathbf{X}),
\end{equation}
the final descriptor satisfies
\[
\text{MPTF-Net}(T_k \text{Input}) 
= 
\text{MPTF-Net}(\text{Input}),
\]
ensuring yaw-rotation invariance without explicit alignment or rotation augmentation.

\subsection{Network Training}
\label{sec:net_training}

We follow the triplet margin loss used in ~\cite{cai2022autoplace} to train the network. Specifically, for each training step, we construct a mini-batch \((q, \ p^q,\ \{n^q_i\})\) containing a query, a positive sample, and several negative samples. A positive sample is defined as one located no more than 9 meters from the query’s capture location, while a negative sample is defined as one located at least 18 meters away.

We denote \( f(\cdot) \) as the mapping from the input to its global descriptor. Our goal is to minimize the distance between \( f(q) \) and \( f(p^q) \), while maximizing the distance between \( f(q) \) and \( f(n^q_i) \). It has been reported that the network is prone to overfitting in the image domain during training [13]. To mitigate this, for each training query, we select the positive sample with the smallest distance between \( f(p^q_i) \) and \( f(q) \) from the potential positive group to form the mini-batch. The negative samples are randomly selected from the potential negative group. Additionally, to increase training efficiency, we only retain negative samples that satisfy \( d(f(q), f(n^q_i)) < \text{margin} \). The loss function is defined as:

\[
L_T = \frac{1}{N_{\text{neg}}} \sum_{i=1}^{N_{\text{neg}}} \left[ d(f(q), f(pq)) - d(f(q), f(nq_i)) + m \right]_+,
\]

where \([ \cdot\cdot \cdot]_+\) denotes the hinge loss, \( d(\cdot) \) is the Euclidean distance, \( m \) is the constant margin, and \( N_{\text{neg}} \) represents the number of selected negative samples.

\section{EXPERIMENT}
\label{sec:experiment}
\begin{table*}[!t]
\centering
\small 
\setlength{\tabcolsep}{0pt} 
\renewcommand{\arraystretch}{1.2} 
\caption{Comprehensive evaluation of place recognition performance on the nuScenes dataset (BS and SON splits).}
\label{tab:overall_results}
\begin{tabular*}{\textwidth}{@{\extracolsep{\fill}}lcccccccc}
\toprule
\multirow{2}{*}{Approach} & \multicolumn{4}{c}{BS split } & \multicolumn{4}{c}{SON split } \\
\cmidrule(lr){2-5} \cmidrule(lr){6-9}
& Recall@1 & Recall@5 & Recall@10 & max $F_1$ & Recall@1 & Recall@5 & Recall@10 & max $F_1$ \\
\midrule
PointNetVLAD~\cite{pointnetvlad} & 74.30 & 84.54 & 87.53 & 0.8885 & 98.49 & 99.43 & 99.58 & 0.9931 \\
Scan Context~\cite{scan} & 85.71 & 93.27 & 97.07 & 0.9324 & 97.80 & 99.54 & 99.37 & 0.9941  \\
MinkLoc3D~\cite{minkloc} & 89.37 & 94.82 & 96.15 & 0.9368 & 98.21 & 99.34 & 99.61 & 0.9917 \\
FusionVLAD~\cite{fusionvald} & —89.21 & 95.61 & 98.10 & 0.9511 & 92.89 & 98.36 & 99.15 & 0.9891 \\
LCPR~\cite{lcpr} & 94.15 & 98.44 & 99.14 & 0.9699 & 99.06 & 99.76 & 99.82 & 0.9953 \\
CVTNet~\cite{cvt} & 94.97 & 98.10 & 99.52 & 0.9914 & 99.11 & 99.69 & 99.91 & 0.9963 \\
\textbf{MPTF-Net (Ours)} & \textbf{96.31} & \textbf{99.00} & \textbf{99.60} & \textbf{0.9813} & \textbf{99.43} & \textbf{99.82} & 99.88 & \textbf{0.9971} \\
\bottomrule
\end{tabular*}
\end{table*}
\begin{table*}[!t]
\centering
\small
\setlength{\tabcolsep}{0pt}
\renewcommand{\arraystretch}{1.2}
\caption{Quantitative comparison of place recognition performance on the NCLT dataset. The best results are highlighted in \textbf{bold}, and the second-best results are \underline{underlined}.}
\label{tab:nclt_comparison}
\begin{tabular*}{\textwidth}{@{\extracolsep{\fill}}lcccccccccccc}
\toprule
\multirow{2}{*}{Approach} & \multicolumn{3}{c}{2012-02-05} & \multicolumn{3}{c}{2013-02-23} & \multicolumn{3}{c}{2013-04-05} & \multicolumn{3}{c}{Mean} \\
\cmidrule(lr){2-4} \cmidrule(lr){5-7} \cmidrule(lr){8-10} \cmidrule(lr){11-13}
 & R@1 & R@5 & R@20 & R@1 & R@5 & R@20 & R@1 & R@5 & R@20 & R@1 & R@5 & R@20 \\
\midrule
PointNetVLAD~\cite{pointnetvlad} & 0.746 & 0.823 & 0.875 & 0.469 & 0.604 & 0.719 & 0.449 & 0.576 & 0.683 & 0.555 & 0.668 & 0.759 \\
Scan Context~\cite{scan}         & 0.767 & 0.836 & 0.909 & 0.481 & 0.564 & 0.726 & 0.418 & 0.496 & 0.649 & 0.555 & 0.632 & 0.761 \\
MinkLoc3D~\cite{minkloc}         & 0.802 & 0.864 & 0.926 & 0.507 & 0.616 & 0.751 & 0.482 & 0.587 & 0.685 & 0.597 & 0.689 & 0.787 \\
FusionVLAD~\cite{fusionvald}     & 0.786 & 0.870 & 0.922 & 0.510 & 0.643 & 0.754 & 0.429 & 0.553 & 0.667 & 0.575 & 0.689 & 0.781 \\
LCPR~\cite{lcpr}                 & 0.856 & 0.891 & 0.935 & 0.669 & 0.798 & 0.830 & 0.651 & 0.712 & 0.802 & 0.725 & 0.800 & 0.856 \\
CVTNet~\cite{cvt}                & \underline{0.924} & \underline{0.937} & \underline{0.951} & \textbf{0.772} & \underline{0.853} & \textbf{0.867} & \textbf{0.780} & \underline{0.826} & \textbf{0.869} & \textbf{0.825} & \underline{0.872} & \textbf{0.896} \\
\textbf{MPTF-Net (Ours)}         & \textbf{0.927} & \textbf{0.941} & \textbf{0.954} & \underline{0.743} & \textbf{0.856} & \underline{0.860} & \underline{0.751} & \textbf{0.832} & \underline{0.856} & \underline{0.807} & \textbf{0.876} & \underline{0.890} \\
\bottomrule
\end{tabular*}
\end{table*}

\begin{table}[!t]
\centering
\small
\setlength{\tabcolsep}{0pt}
\renewcommand{\arraystretch}{1.15}
\caption{Quantitative comparison of place recognition performance on the KITTI dataset. The best results are highlighted in \textbf{bold}.}
\label{tab:kitti_results}
\begin{tabular*}{\columnwidth}{@{\extracolsep{\fill}}l l c c}
\toprule
Dataset & Approach & AUC & max $F_1$ \\ 
\midrule
\multirow{7}{*}{KITTI}
& PointNetVLAD~\cite{pointnetvlad} & 0.856 & 0.846 \\
& Scan Context~\cite{scan}         & 0.836 & 0.835 \\
& MinkLoc3D~\cite{minkloc}    & 0.894 & 0.869 \\
& FusionVLAD~\cite{fusionvald}     & 0.871 & 0.855 \\
& LCPR~\cite{lcpr}                 & .862 & 0.854 \\
& CVTNet~\cite{cvt}                & 0.891 & 0.879 \\
& \textbf{MPTF-Net (Ours)}         & \textbf{0.897} & \textbf{0.883} \\
\bottomrule
\end{tabular*}
\end{table}
\subsection{Dataset}

\textbf{nuScenes Dataset:} Containing 1000 scenarios from Boston and Singapore, nuScenes~\cite{nuscenes} is utilized to assess cross-domain generalization. We train on the Boston Seaport (BS) split and evaluate on the unseen Singapore One-North (SON) split for zero-shot assessment, strictly following the protocol in~\cite{lcpr}.

\textbf{NCLT Dataset:} To evaluate long-term robustness, we employ the NCLT Dataset~\cite{nclt}, capturing 27 sessions over 15 months with significant seasonal variations. We leverage its repetitive trajectories to benchmark resilience against drastic environmental changes, including dynamic occlusions, foliage shifts, and snow cover.

\textbf{KITTI Dataset:} To verify system versatility, we employ the KITTI odometry benchmark~\cite{kitti}, which features calibrated LiDAR sequences in diverse environments. These trajectories allow for rigorous evaluation of loop closure performance under distinct sensor specifications.
\subsection{Implementation Details and Evaluation Metrics}

\textbf{Input Settings:} For the nuScenes dataset (BS and SON splits) and NCLT dataset, the LiDAR BEV input size is set to $4 \times 32 \times 1056$, and the RIV input size is $2 \times 32 \times 1056$. This configuration balances feature resolution with computational efficiency. For the KITTI dataset, we utilize a higher vertical resolution, setting the LiDAR BEV and RIV input sizes to $4 \times 64 \times 1056$ and $2 \times 64 \times 1056$, respectively.

\textbf{Training Settings:} For the nuScenes dataset, following LCPR~\cite{lcpr}, positive samples are defined as point clouds within 9 meters of the query anchor, while negative samples are those beyond 18 meters. For the KITTI and NCLT datasets, positive samples are defined by an overlap ratio greater than 0.3, with negatives falling below this threshold. The NetVLAD layer is initialized randomly and jointly optimized with the backbones. We employ the Adam optimizer with an initial learning rate of $1\times10^{-5}$, decaying by a factor of 10 every 10 epochs. All experiments are conducted on a single NVIDIA RTX 4090 GPU with 24GB VRAM. 

\textbf{Evaluation Metrics:} We report Recall@$k$ ($k=1, 5, 10$) and the maximum $F_1$ score to evaluate retrieval accuracy and robustness. Recall@1 indicates immediate localization capability, while max $F_1$ provides a balanced view of precision and recall. Additionally, following standard benchmarks, we report the Area Under the Curve (AUC) of the precision-recall curve to assess the global effectiveness of the system.

\subsection{Evaluation for Place Recognition}

To rigorously benchmark the generalization and robustness of MPTF-Net, we conducted evaluations across three diverse datasets. Following the protocol in~\cite{lcpr}, we first utilized the nuScenes dataset to assess zero-shot generalization, training models on the Boston Seaport (BS) split and evaluating on both the seen BS and unseen Singapore One-North (SON) splits. To further verify robustness against long-term environmental changes, we employed the NCLT dataset, selecting the 2012-01-08 session for training and the 2012-02-05, 2013-02-23, and 2013-04-05 sessions for testing. Additionally, the KITTI odometry benchmark (sequences 03--10 for training, sequence 00 for evaluation) was used to confirm system versatility across varying sensor specifications.

Quantitative assessments reported in Table~\ref{tab:overall_results}, Table~\ref{tab:nclt_comparison}, and Table~\ref{tab:kitti_results} substantiate that MPTF-Net establishes a new state-of-the-art across all testing scenarios. On the nuScenes benchmark, our method demonstrates exceptional zero-shot resilience, outperforming the multimodal baseline LCPR by 2.16\% on the seen split and extending the lead over AutoPlace to 6.44\% on the unseen split. This transferability stems from our absolute-position-free Transformer, which learns invariant geometric fingerprints rather than overfitting to specific spatial layouts. Extending this robustness to the temporal domain, MPTF-Net maintains high stability on the NCLT dataset despite severe seasonal variations, achieving the highest Recall@5 across all test sessions. This validates that our NDT-enriched BEV representation provides structural priors that remain distinct even when visual textures degrade. Furthermore, comparisons on KITTI confirm the system's versatility, where MPTF-Net exceeds the leading sparse-convolution method MinkLoc3D by 1.3\% in AUC. By hierarchically fusing fine-grained local textures with global structural dependencies, our single-frame approach overcomes the receptive field limitations of conventional CNNs, delivering an optimal precision-recall trade-off.

\subsection{Ablation Experiment}

To provide a comprehensive analysis of MPTF-Net, we conducted two sets of ablation studies to isolate the contributions of specific architectural components and validate the optimal configuration of the multi-scale fusion strategy.

\noindent\textbf{Impact of Key Modules.} The first study investigates the effectiveness of the proposed NDT-based geometric representation and the necessity of the Cross-Attention mechanism. Results are presented in Table~\ref{tab:ablation_bs_f1}. Replacing the NDT-BEV features with standard BEV features (based on simple statistical aggregation) causes a Recall@1 drop from 96.31\% to 94.21\%. This degradation highlights that traditional statistical operations fail to capture fine-grained geometric structural information, which is effectively preserved by our NDT encoding. Furthermore, replacing Cross-Attention with standard Self-Attention results in a significant drop to 90.22\%. This decline underscores the insufficiency of independent modality processing; cross-modal interaction is indispensable for synthesizing discriminative descriptors that leverage complementary cues from both RIV and BEV streams.

\begin{table}[t]
\centering
\caption{Ablation study of proposed MPTF-Net on the nuScenes dataset BS split.}
\label{tab:ablation_bs_f1}
\setlength{\tabcolsep}{3pt}
\renewcommand{\arraystretch}{1.1}

\begin{tabular*}{\columnwidth}{@{\extracolsep{\fill}}lcccc}
\toprule
\multirow{2}{*}{Approach} & \multicolumn{4}{c}{BS Split} \\ 
\cmidrule(lr){2-5}
& R@1 & R@5 & R@10 & Max $F_1$ \\
\midrule
\textbf{Ours (w/ NDT-BEV)} & \textbf{96.31} & \textbf{99.00} & \textbf{99.60} & \textbf{0.9813} \\
\midrule
w/ Std. BEV  & 94.21 & 98.40 & 99.00 & 0.9671 \\ 
w/ Two-scale Fusion & 93.61 & 94.01 & 97.14 & 0.9417 \\
w/ Self-Attn & 90.22 & 97.60 & 98.30 & 0.9353 \\ 
\bottomrule
\end{tabular*}
\end{table}

\begin{figure}[t]
\centering %
\includegraphics[width=0.48\textwidth]{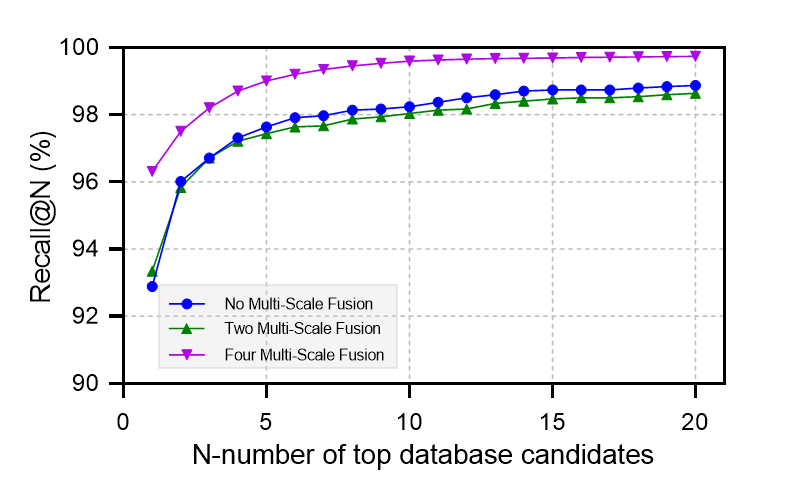} %
\caption{Analysis of multi-Scale fusion strategies on recall performance.}
\label{fig:multi}
\end{figure}

\noindent\textbf{Effectiveness of Multi-Scale Fusion.} The second study analyzes the impact of fusion granularity within the pyramid architecture. We evaluate performance under three configurations: single scale (no fusion), two-scale fusion, and the proposed four-scale fusion. The Recall@$N$ curves are plotted in Fig.~\ref{fig:multi}. Results indicate a distinct performance hierarchy. While the two-scale configuration offers only marginal improvement over the non-fused baseline, the four-scale approach demonstrates a decisive advantage, boosting Recall@1 to 96.31\%. This confirms that deep hierarchical aggregation is essential for capturing both fine-grained local textures and high-level semantic context. Crucially, the four-scale design represents an optimal trade-off between retrieval accuracy and computational complexity, avoiding the diminishing returns and excessive parameter overhead associated with deeper architectures.

\subsection{Runtime and Memory Consumption}

\begin{figure}[t]
\centering
\includegraphics[width=0.48\textwidth, trim={0 0 2pt 0}, clip]{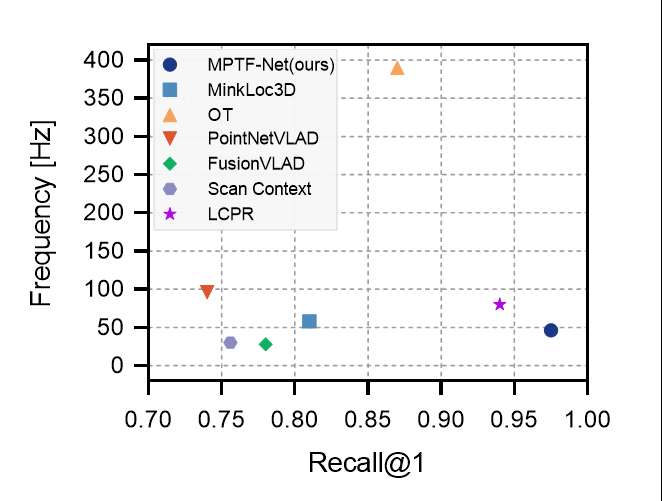}
\caption{Comparison of runtime and efficiency with state-of-the-art methods.}
\label{fig:runtime}
\end{figure}

We evaluate the runtime efficiency and memory footprint of MPTF-Net on the nuScenes dataset under the experimental settings described in Sec. IV-B. For each query, we measure the latency of descriptor generation and top-20 retrieval. As shown in Fig. \ref{fig:runtime}, on the BS split with 9,686 database samples, MPTF-Net achieves an average end-to-end runtime of 19.62 ms with 35.83M parameters, including 11.70 ms for descriptor generation and 7.92 ms for retrieval, demonstrating its real-time capability for online autonomous driving.
\subsection{Yaw-Rotation Invariance Study}

\begin{figure}[t]
\centering
\includegraphics[width=0.48\textwidth]{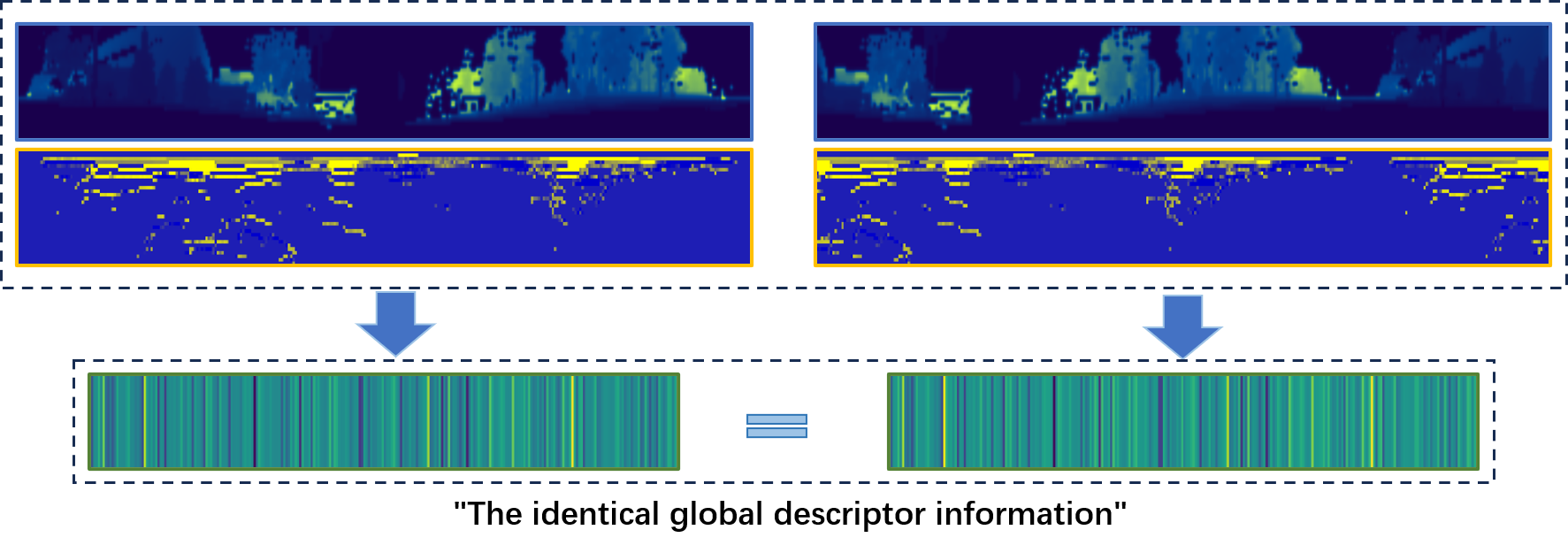}
\caption{Visual validation of rotation invariance. The upper left shows the original input; upper right displays the input after a 55$^\circ$ yaw rotation. The lower row presents the corresponding global descriptors generated by MPTF-Net, showing high structural consistency.}
\label{fig:yaw_vis} 
\end{figure}

The yaw-rotation invariance of MPTF-Net, theoretically formulated in Section~\ref{sec:yaw_invariance}, is further corroborated through extensive empirical validation on the BS dataset split. As illustrated in Fig.~\ref{fig:yaw_vis}, the network jointly processes two complementary modalities—RIV and BEV—capturing both fine-grained geometric cues and global spatial context. To systematically evaluate rotational robustness, a representative frame was rotated by yaw angles of 55$^\circ$, 110$^\circ$, 180$^\circ$, 250$^\circ$, 305$^\circ$, and 360$^\circ$. Visualization results show that the structural patterns of the extracted descriptors remain strikingly consistent across all rotations. This provides strong empirical evidence that MPTF-Net not only achieves theoretical invariance but also retains stable representations under substantial viewpoint changes.

\begin{figure}[t]
\centering
\includegraphics[width=0.48\textwidth]{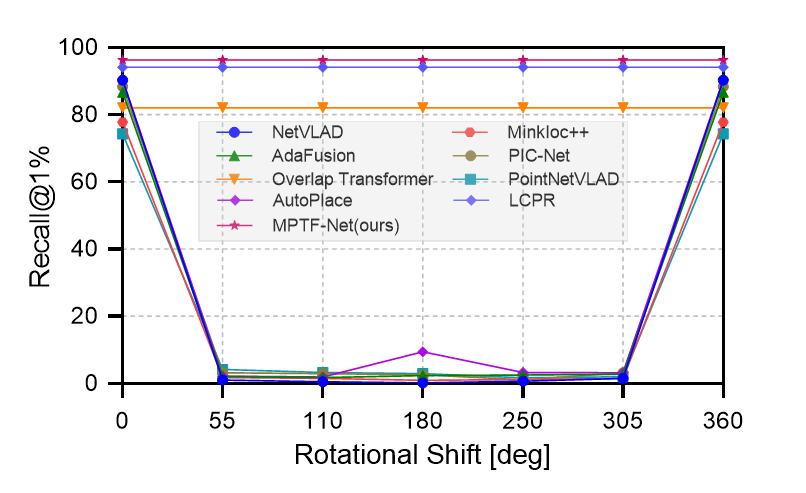}
\caption{Quantitative study on yaw-rotation invariance comparing Recall@1 across different rotation angles.}
\label{fig:yaw_quant} 
\end{figure}

To quantitatively assess this robustness, Recall@1 was employed as the primary metric. As shown in Fig.~\ref{fig:yaw_quant}, MPTF-Net, LCPR, and OverlapTransformer demonstrate clear resilience to axial rotations, whereas several competing methods exhibit significant performance degradation. Although LCPR and OverlapTransformer also maintain rotation-invariant behavior, MPTF-Net attains consistently higher Recall@1 scores, highlighting its enhanced capacity to preserve discriminative information under rotational perturbations. These results underscore the effectiveness of MPTF-Net in learning rotation-stable descriptors.

\section{CONCLUSIONS}
In this study, we propose MPTF-Net, a novel multi-view, multi-scale fusion network for LiDAR-based place recognition. At the input level, our method leverages multi-channel RIV and BEV representations generated via Normal Distribution Transform, where the BEV encodes local point cloud statistics to enhance robustness against noise. At the architecture level, the Transformer-based network captures complementary correlations across multiple views and integrates multi-scale contextual features. Extensive experiments on multiple datasets demonstrate that MPTF-Net produces highly discriminative global descriptors, surpassing both single-view and multi-view baselines, while maintaining real-time performance suitable for practical deployment in online applications.

\bibliographystyle{IEEEtran}
\bibliography{ref}

\end{document}